\documentclass[letter``, 10pt, conference]{ieeeconf}      

\IEEEoverridecommandlockouts                              

\usepackage{graphicx}
\usepackage{amsmath}
\usepackage{amssymb}
\usepackage{subcaption}
\usepackage{hyperref}
\usepackage{draftwatermark}

\usepackage{array,multirow,graphicx}

\graphicspath{ {imgs/} }
\usepackage{tikz}

\usepackage[final, markup=bfit, deletedmarkup=sout, authormarkup=brackets]{changes}
\colorlet{Changes@Color}{blue}
\definechangesauthor[name={Petr}, color=red]{PS}
\definechangesauthor[name={Michael}, color=pink]{MT}
\definechangesauthor[name={Matej}, color=orange]{MH}
\definechangesauthor[name={Tomas}, color=green]{TS}
\definechangesauthor[name={Karla}, color=purple]{KS}
\definechangesauthor[name={Jan}, color=cyan]{JB}

\begin{document}
\SetWatermarkAngle{0}
\SetWatermarkColor{black}
\SetWatermarkLightness{0.5}
\SetWatermarkFontSize{10pt}
\SetWatermarkVerCenter{20pt}
\SetWatermarkText{\parbox{30cm}{%
\centering This is the author final version of the manuscript published in\\
\centering 2019 IEEE/RSJ International Conference on Intelligent Robots and Systems (IROS), pp.  7574-7581\\
\centering Macau, China, November 4-8, 2019; https://doi.org/10.1109/IROS40897.2019.8968463, (C) IEEE.}}

%
\title{\LARGE \bf Safe physical HRI: Toward a unified treatment of speed and separation monitoring together with power and force limiting}
%
%
%

\author{Petr~Svarny,
        Michael~Tesar,
        Jan~Kristof~Behrens,
        and~Matej~Hoffmann
\thanks{Petr Svarny and Matej Hoffmann are with the Department of Cybernetics, Faculty of Electrical Engineering, Czech Technical University in Prague. Michael Tesar and Jan K. Behrens with the Czech Institute of Informatics, Robotics, and Cybernetics of the Czech Technical University in Prague. (e-mail: petr.svarny@fel.cvut.cz; michael.tesar@cvut.cz; jan.kristof.behrens@cvut.cz, matej.hoffmann@fel.cvut.cz).}
}

\maketitle

\begin{abstract}

So-called collaborative robots are a current trend in industrial robotics. However, they still face many problems in practical application such as reduced speed to ascertain their collaborativeness. The standards prescribe two regimes: (i) speed and separation monitoring and (ii) power and force limiting, where the former requires reliable estimation of distances between the robot and human body parts and the latter imposes constraints on the energy absorbed during collisions prior to robot stopping. Following the standards, we deploy the two collaborative regimes in a single application and study the performance in a mock collaborative task under the individual regimes, including transitions between them. Additionally, we compare the performance under ``safety zone monitoring'' with keypoint pair-wise separation distance assessment relying on an RGB-D sensor and skeleton extraction algorithm to track human body parts in the workspace. Best performance has been achieved in the following setting: robot operates at full speed until a distance threshold between any robot and human body part is crossed; then, reduced robot speed per power and force limiting is triggered. Robot is halted only when the operator's head crosses a predefined distance from selected robot parts. We demonstrate our methodology on a setup combining a KUKA LBR iiwa robot, Intel RealSense RGB-D sensor and OpenPose for human pose estimation. 
\end{abstract}



%

\IEEEpeerreviewmaketitle

\section{Introduction}

So-called ``collaborative robots'' (or ``cobots''), i.e. robots that are safe when sharing the same (collaborative) workspace with human operators, represent a rising trend in robotics. However, their industrial application is limited by their performance---the reduced speed and limited payload in particular.
 Safe \textit{physical Human-Robot Interaction} (pHRI) saw great development in the last decade, with the introduction of new safety standards \cite{ISO10218,ISO/TS15066} and a rapidly growing market of cobots. However, it is a more recent attempt to enhance not only the safety of these robots but also their performance. This attempt to make collaborative robotics more attractive to the traditional industry is visible also in projects promoting the advancement in this field (see the COVR project\footnote{\url{http://safearoundrobots.com/}} \cite{Bessler2019}).
 
 Haddadin and Croft \cite{Haddadin2016} provide a survey of pHRI. According to \cite{ISO/TS15066}, there are two ways of satisfying the safety requirements when a human physically collaborates with a robot: (i) \textit{Power and force limiting (PFL)} and (ii) \textit{Speed and separation monitoring (SSM)}. For PFL, physical contacts with a moving robot are allowed but the forces / pressures / energy absorbed during a collision need to be within human body part specific limits. This translates onto lightweight structure, soft padding, no pinch points, and possibly introduction of elastic elements (see the series elastic actuators in Sawyer robot; \cite{Haddadin2017} for a formal treatment of robots with flexible joints) on the robot side, in combination with collision detection and response relying on motor load measurements, force/torque or joint torque sensing. This is addressed by interaction control methods for this \textit{post-impact} phase (see \cite{Haddadin2017} for a recent survey). The performance of robots complying with this safety requirement in terms of payload, speed, and repeatability is limited.  

Safe collaborative operation according to speed and separation monitoring prohibits contacts with a moving robot and thus focuses on the \textit{pre-impact} phase: a \textit{protective separation distance}, $S_p$, between the operator and robot needs to be maintained at all times. When the distance decreases below $S_p$, the robot stops \cite{ISO/TS15066}.

In industry, $S_p$ is typically safeguarded using light curtains (essentially electronic versions of physical fences) or safety-rated  scanners that monitor 2D or 3D zones (e.g., Pilz SafetyEYE). One can usually define a protection field (denoted ``red'' zone)---if an object is detected inside, the robot is brought to an immediate halt---and a warning field (called ``yellow'' zone) that may trigger a reduced maximum allowed robot speed. However, the flexibility of such setups is limited: the information is reduced to detecting whether an object of a certain minimum volume has entered one of the two predefined zones. Also, the higher the robot kinetic energy, the bigger is its footprint on the shop floor. 

\begin{figure}
	\centering
	\begin{subfigure}[b]{0.52\columnwidth}            
	    \includegraphics[width=\columnwidth]{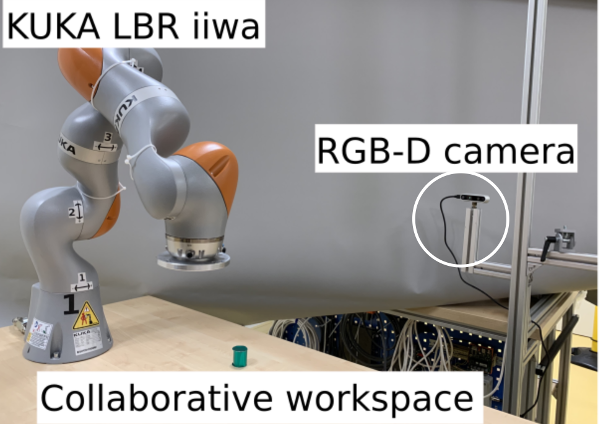}
            \caption{}
            \label{fig:ws}
    \end{subfigure}%
    \begin{subfigure}[b]{0.48\columnwidth}
            \centering
            \includegraphics[width=\columnwidth]{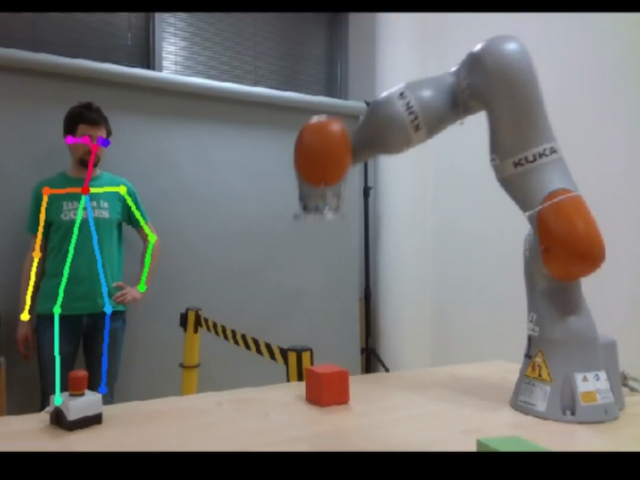}
            \caption{}
            \label{fig:camera_view}
    \end{subfigure}
	
	\caption{Experimental setup -- collaborative workspace. (a) External view. (b) Camera view with human keypoint extraction.}
	\label{fig:workspace}
\end{figure} 

With increasing performance and falling prices of RGB-D sensors (RGB image + depth information), we can prototype collaborative scenarios using already available sensors (like Intel RealSense) and tools for human keypoint or skeleton extraction from camera images \cite{Cao2017,Insafutdinov2016}. This combination permits real-time perception of the positions of individual body parts of any operators in the collaborative workspace. Deployment in real applications will depend on the development of safety-rated modules providing this functionality\footnote{For example , \url{https://www.veobot.com/}}. 

In this work, we take advantage of the keypoint information and follow \cite{ISO/TS15066} to deploy the two collaborative regimes (SSM and PFL) in a single application. 
The deployment of both regimes in a single scenario provides in our view the unique contribution of this work. The PFL regime prescribes different thresholds for the body parts of the operator and hence only with the keypoint information available can the body part specific limits be taken into consideration---demonstrated on the head keypoints here. We study the performance in a mock collaborative task under different settings like distances from robot base to individual keypoints, stopping or slowing down, and their transitions---the distances and speeds are based on  \cite{ISO/TS15066} in our setup.
We use a KUKA LBR iiwa collaborative robot, Intel RealSense RGB-D sensor and OpenPose for human pose estimation as shown in Fig. \ref{fig:workspace}.


This article is structured into related work reviewed in the next section, followed by Materials and Methods, and Results. We close by Discussions and Conclusions. 

 \section{Related work}
 \label{sec:related}
A functional solution for safe pHRI according to the speed and separation monitoring requirements will necessarily involve: (i) sensing of the human operators' as well as robot's positions (and speeds), (ii) a suitable representation of the corresponding separation distances, and (iii) appropriate responses of the machine (speed reduction / stop / avoidance maneuvers). On the perception side, tracking the robot parts in space tends to be relatively easy as accurate models of the machine as well as joint encoder readings are available and hence position (and possibly also orientation, speed, and acceleration) for the end-effector as well as other chosen keypoints can be readily obtained from forward kinematics. On the other hand, the perception of the human operators in the workspace is more challenging. 
Two key technologies have appeared that facilitate progress in this area: (i) compact and affordable RGB-D sensors and (ii) convolutional neural networks for human keypoint/skeleton extraction from camera images \cite{Cao2017,Insafutdinov2016}, or full 3D human body reconstruction \cite{Guler2018}. These technologies together---albeit currently not safety-rated---make it possible to perceive the positions of individual body parts of any operators in the collaborative workspace in real time. Alternative technologies include distributed wireless sensor networks that track operators who do not wear any devices \cite{Savazzi2016} or proximity sensors distributed on the robot, usually part of electronic skins (e.g., Bosch APAS robot). The main benefit of all these solutions is their resolution---compared to mere zone monitoring---and hence reduction of the effective footprint of the robot.

Once the robot and human positions are obtained, their relative distances (and possibly speeds or time to collision) need to be evaluated. Euclidean distance is the most natural candidate and also one that appears in the safety norms. However, other representations have been proposed and may be better suited for the nature of the sensory data (like the depth space approach for RGB-D data \cite{Flacco2012,Flacco2015}) or for planning and control of the robot where the configuration space (joint space) of the robot can be used for representing both the robot body and the obstacles. Flacco et al.~\cite{Flacco2015} provide an overview. Another key component is in what form are the robot and human body parts represented. Drawing on the results of the computer graphics community (\cite{Jimenez2001} for a survey), this often takes the form of some collision primitives. These can be simple shapes like spheres \cite{Flacco2012} or more complex meshes \cite{Polverini2017} and can differ for the robot and the human: Zanchettin et al.~\cite{Zanchettin2016} represent robot links as segments and humans as a set of capsules. Of course, for safety to be guaranteed, the whole body of both agents should be represented and considering only the robot end-effector does not suffice. Often, the ``robot-centered'' approach is taken---in the sense that the collision primitives are centered on the robot body and possibly dynamically shaped based on the current robot velocity \cite{Polverini2017,Zanchettin2016,Lacevic2010,Magnanimo2016}. A biologically inspired approach relying on peripersonal space representation was presented in \cite{Roncone2016,Nguyen2018}.

Interaction control methods for the \textit{post-impact} phase (see \cite{Haddadin2017} for a survey) are not our focus here. 
We rely mainly on the information in \cite{ISO/TS15066} to calculate the speed our robot can run with while fulfilling the PFL regime criteria.

There is a large body of work dealing with motion planning and control in dynamic environments. In the face of dynamically appearing obstacles (the case in HRI scenarios), classical offline trajectory planning \cite{LaValle2001} has to be complemented by reactive strategies \cite{Khatib1986,Brock2002}. This problem gives rise to new velocity-dependent formulations such as ``velocity obstacles'' \cite{Fiorini1998} or ``dynamic envelope'' \cite{Vatcha2009}. Recently, the approaches are somewhat closer to the ``control'' than to the ``planning'' community: the work of De Luca and Flacco (\cite{Flacco2012}; \cite{DeLuca2012} deal with both pre-impact and post-impact control) or Zanchettin et al. \cite{Zanchettin2016} are good examples. 
In summary, researchers in robotics often find themselves developing compelling solutions for real-time obstacle avoidance, but these may require substantial tuning and the separation distance is often optimized rather than guaranteed (e.g., \cite{Lacevic2010, Khatib1986}). There are notable exceptions like the work of Marvel \cite{Marvel2013} and Zanchettin et al.~\cite{Zanchettin2016} that take the constraints imposed by the safety standards seriously. Regarding the PFL regime, Sloth and Petersen~\cite{Sloth2018} recently presented a method to compute safe path velocities complying with \cite{ISO/TS15066}; Mansfeld et. al.~\cite{Mansfeld2018} developed a ``safety map'' and use alternative, less conservative, collision limits derived from biomechanics impact data. Similarly, \cite{Meguenani2015,Rossi2015} provide a treatment of robot control taking into account the energy dissipated in possible contacts with the operator.

The SSM part of our framework follows up on our previous  work \cite{Nguyen2018, Svarny_SSR_2018}, in which we take advantage of the keypoint extraction to monitor distances between individual parts of the human and robot body and exploit also the keypoint semantics to modulate the behavior.
In this work, we make important steps in bringing these ideas to an industrial setting by moving to an industrial collaborative robot, adding the PFL regime, and illustrating how to determine all the relevant parameters in accordance with \cite{ISO/TS15066}.

\section{Materials and Methods}


\subsection{Robot platform}
\label{subsec:robot_control}
A 7 DoF industrial manipulator KUKA LBR iiwa 7 R800 was used. The robot operates either at full speed (up to 1~m/s for the end-effector) or reduced speed (0.42~m/s). As an additional low-level safety layer, the KUKA \textit{Collision detection} based on external torque estimation was turned on.

\subsection{RGB-D camera}
\label{subsec:camera}
The camera was an Intel RealSense D435 RGB-D. We calibrate the robot and camera position through the ROS Hand-Eye calibration tool. The camera resolution is 848x480, and we use the RealSense short range presets\footnote{See the file ShortRangePreset.json in the wiki pages at \cite{Intel2018}.}.

\subsection{HRI setup}
Our setup is illustrated in Fig.~\ref{fig:workspace}. A mock collaborative task has been staged: the robot performs a periodic operation. Operator periodically replaces one of the objects, entering the robot workspace, and is perceived by the camera. The robot responds appropriately (slow down or stop). The robot was placed on a fixed table while the RGB-D sensor was on a fixed position so that it can capture the whole robot workspace. The camera was fixed to a construction that was separate from the robot's platform to avoid tremors during the robot's movement. The setup was designed to minimize the chance of occlusions.\footnote{The complete setup including all experimental scenarios is illustrated in the accompanying video at \url{https://youtu.be/zP3c7Eq8yVk}.}

\subsection{Software framework and robot control}
\label{subsec:software_framework}

\begin{figure}
	\centering
	\includegraphics[width=3in]{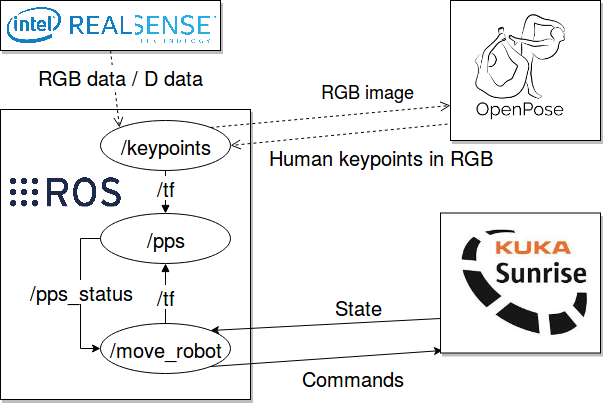}
	\caption{Software architecture schematics.}
	\label{fig:modules}
\end{figure}

A schematics of the overall framework is shown in Fig.~\ref{fig:modules}. 
OpenPose (see Sec.~\ref{subsec:openpose}) finds human keypoints in pictures captured by the camera as orchestrated by a ROS node. The robot node consumes and produces information about the coordinate transformations. The relative distances are assessed in the peripersonal space module (\textit{pps}) and fed into the robot controller to generate the appropriate response.

High-level control of the robot was done in the ROS node \textit{move\_robot}. 
We used the MoveIt! motion planning framework \cite{Coleman2014} to generate and execute the trajectories for our mock task. Our scenario additionally required speed modulation (stop, slow down, speed up) on the run which is not provided by Moveit! and we have implemented a custom solution for smoothly modulating the trajectories in joint space, compliant with the corresponding limits of the platform. 
In brief, we used cascaded robot control which masks system non-linearities and lets us see the robot as a system of seven double-integrators, which we control similarly to a saturation controller \cite{Rao2001}.
We distinguish:

\noindent(i) \textbf{Stopping motion.} The remaining trajectory of the robot is replaced by an alternative trajectory with a maximal deceleration for the fastest joint and relatively scaled deceleration for all other joints. The overall stopping time $t_e$ is dependent on the velocity of the joints $\Dot{x}_j$ and the acceleration limits $a_\mathrm{j,min} \leq \ddot{x}_j \leq a_\mathrm{j,max}$, $t_{\mathrm{stop},j}$ denotes the minimal stopping time for a joint $j$:

\begin{align}
t_e &= \max_{j \in Joints} t_{\mathrm{stop},j}  \\ 
t_{\mathrm{stop},j} &=  \begin{cases} 
      \Dot{x}_j < 0 & \frac{0 - \Dot{x}_\mathrm{j,ref}}{a_\mathrm{j,max}}  \\
      \Dot{x}_j \geq 0 & \frac{0 - \Dot{x}_\mathrm{j,ref}}{a_\mathrm{j,min}}
   \end{cases}
\end{align}

The worst-case run-time of the stopping trajectory calculation $t_\mathrm{calc} \leq 0.02\ s$ was determined empirically. When the stop signal arrives, the earliest future state (with $t \geq t_\mathrm{now} + t_\mathrm{calc}$) along the current trajectory is selected and used as reference state $\mathbf{x}_\mathrm{ref}$ for calculations.


\begin{align}
\Dot{x}_j(0) &= b_{j,1} = \Dot{x}_\mathrm{j,ref}\\
\Dot{x}_j(t_e) &= b_{j,1} + b_{j,2} t_e = 0 \longrightarrow b_{j,2} = \frac{\Dot{x}_\mathrm{j,ref}}{2t_e}\\
x_j(t) &= \underbrace{x_\mathrm{j,ref}}_{b_{j,0}} + \underbrace{\Dot{x}_\mathrm{j,ref}}_{b_{j,1}} t + \underbrace{\frac{\Dot{x}_\mathrm{j,ref}}{2t_e}}_{b_{j,2}} t^2 
\end{align}

\begin{figure}
	\centering
	\includegraphics[width=0.5\textwidth]{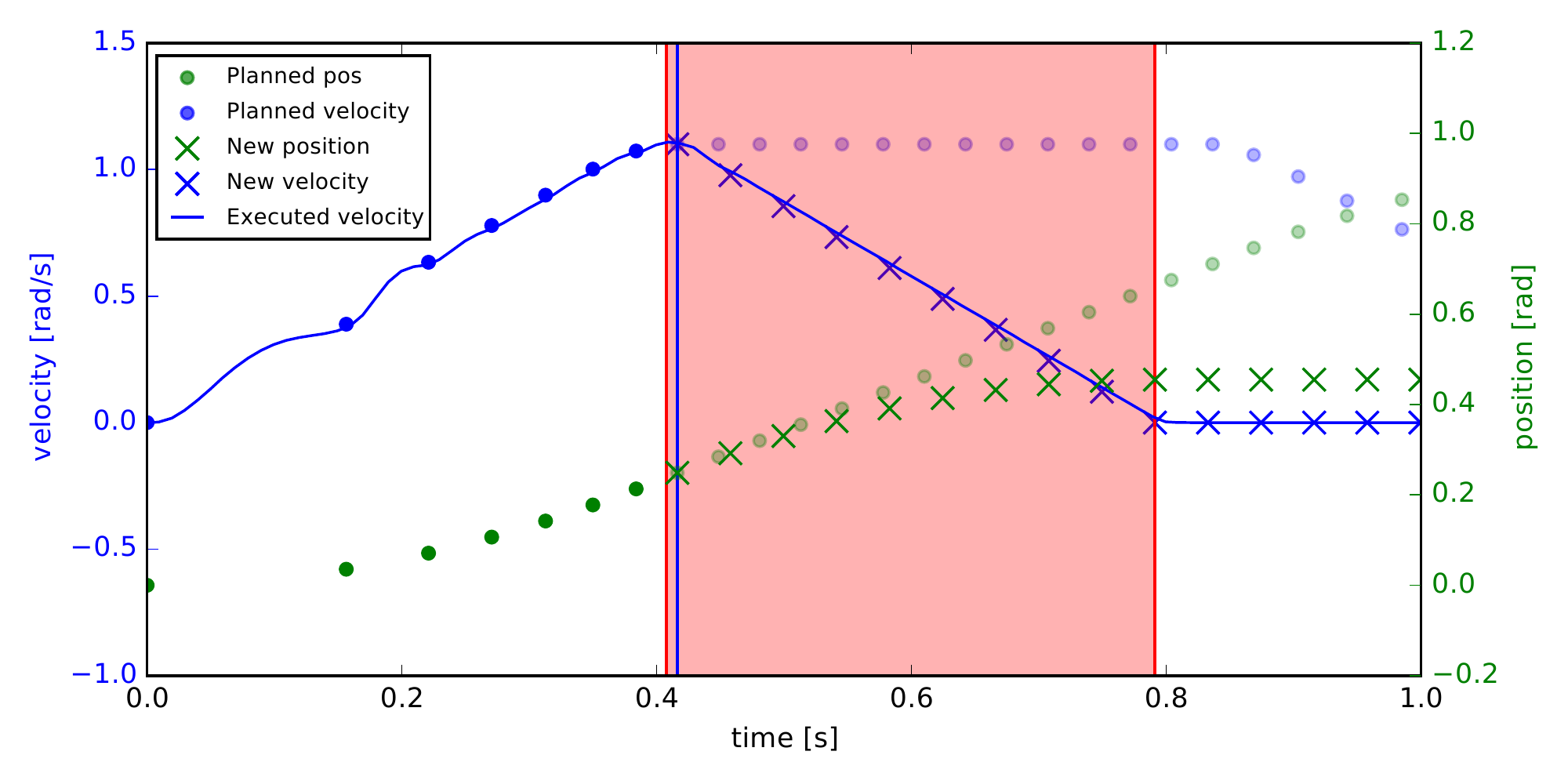}
	\caption{Stopping motion using the trajectory controller. 
	The stopping signal was received at $0.4077$ s. The deceleration starts $0.0086$~s later. The robot stops from the a speed of 1.1~rad /s in $0.3836$~s (red area). The first red vertical line shows arrival of stop signal and the blue vertical line ($0.0085$~s later) marks the end of computation of the new trajectory. Note, that we consider the worst-case execution time in the selection of the reference state. }
	\label{fig:stoppingtraj}
\end{figure}

To facilitate the full breaking potential, we use polynomials (with parameters $b_{j,0}$, $b_{j,1}$ and $b_{j,2}$) of degree two to describe the joint positions. Hence, the velocities $\Dot{x}_\mathrm{j}$
 are linear with the maximum deceleration for at least one joint. This breaking behavior yields the shortest stopping time possible, but will for general trajectories slightly deviate from the original path. For point-to-point movements in free space (as in our example), this stopping strategy will remain on the planned path. Figure~\ref{fig:stoppingtraj} shows the planned joint velocity and position, the stopping plan, and the joint velocity of a simulated robot.
 
 \noindent(ii) \textbf{Deceleration to reduced speed.} 
 When the signal arrives to slow down, a stopping trajectory is calculated as above. The original trajectory is scaled using the \textit{IterativeParabolicTimeParameterization} (MoveIt!) to comply with the desired reduced speed. When the linear deceleration reaches the speed of the scaled trajectory, we search for the closest trajectory point ahead of the scaled trajectory. The scaled trajectory is shifted in time to continue after the deceleration and both trajectories are stitched at this point together. Acceleration back to full speed is performed similarly.
 

The target joint position commands were then passed to the KUKA Sunrise cabinet via the FRI interface.

We took a conservative approach in the design of our controller as follows: when ``pps status'' signaled a more restrictive regime, it was executed immediately; conversely, in the other direction, a filter was applied to warrant that the operator has left the area. The  pipeline described above is not safety-rated and the high-level robot control is capable of performing a Stop Category 2 only.

\subsection{Human keypoint 3D estimation and distance measurements}
\label{subsec:openpose}

The integral part of collision avoidance is to correctly estimate the position of the operator's keypoints in space. We created a ROS node that processed data from the Realsense D435 camera using the Realsense Python API (2.17.1)\cite{Intel2018} to collect aligned color and depth images. All our image operations also rely on OpenCV3\cite{Bradski2000}.

The color images were sent to the OpenPose library Python API\cite{Cao2018} to estimate human keypoints. For OpenPose, we use the COCO model and with the net resolution matching the input images. We also used the model's confidence value to drop detections that were below 0.6 confidence as they were often false positives. This threshold was found by letting OpenPose analyze a scene without the human.


The resulting keypoint locations were then deprojected using the aligned depth image and thus we received the 3D coordinates of the operator in the camera's frame of reference. These keypoints are represented as reference frames and added to the ROS transform library (called \emph{tf}). The \textit{tf} package stores the relationships between different coordinate frames in a tree structure, allowing for calculation of the position of the human keypoints w.r.t. the robot's keypoints by using the relation between their frames.

Our experiment takes into account only upper body and hip keypoints detected by OpenPose's posture model (see Fig.~\ref{fig:bounding}b), namely keypoints 0--7 and 14--17. These are the most relevant keypoints to our application and assume standard behavior of the operator. What we consider for our experiment as the human \emph{head} are the keypoints of the nose (0), eyes (14, 15) and ears (16, 17). 

\subsection{Keypoint ``bounding spheres''}
\label{subsec:bounding_spheres}
Discrete keypoints allow a faster calculation of distances and unambiguous interpretability of the system's expected behavior. Nevertheless, they do not take into account the full occupancy of the bodies, which could lead to the underestimation of the real separation distance. This problem is especially relevant with sparsely placed keypoints.

\begin{figure}
	\centering
            \includegraphics[width=2.5in]{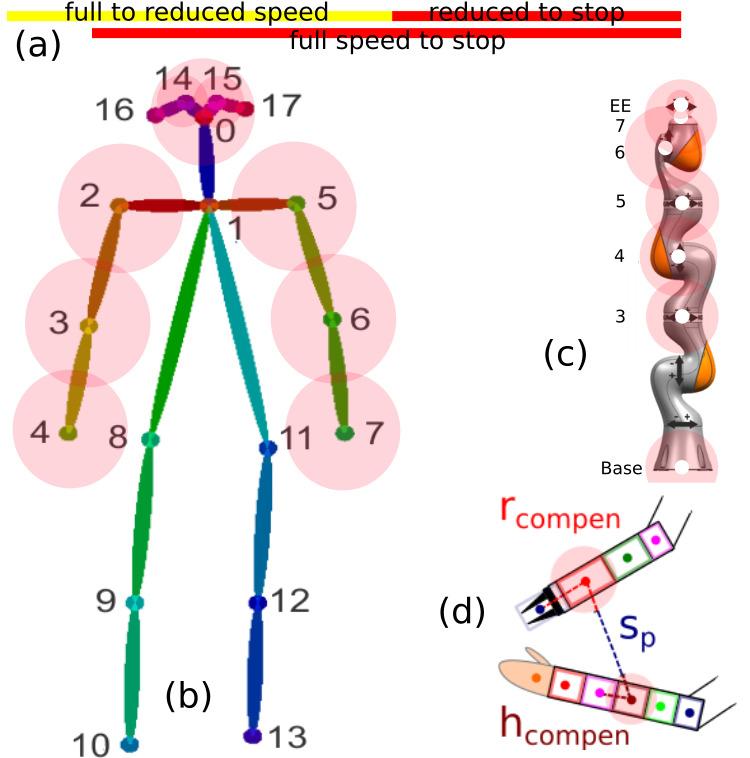}
	\caption{Keypoints and bounding spheres representation (aspect ratio kept). (a) Stopping and stopping after reduced speed distances. (b) OpenPose keypoint distribution~\cite{Cao2017} with bounding spheres on the keypoints of interest. (c) KUKA LBR iiwa keypoints (picture source: KUKA LBR iiwa brochure) with compensation bounding spheres. (d) Schematic 2D separation distance calculation between robot and human keypoints. The compensation coefficients are the distances between the keypoints and the farthest point of the body that belongs to the body part near the keypoint.}
	\label{fig:bounding}
\end{figure}

We need to guarantee $S_{\mathrm{p}}$, the protective separation distance \cite{ISO/TS15066}. For this purpose, we introduce compensation coefficients for the robot $r_{\mathrm{compen}}$ and the human $h_{\mathrm{compen}}$.

The calculation of the compensation coefficients with given keypoints is divided into two steps. In the first step, every part of the body is assigned to its nearest keypoint. Then, for every keypoint, the maximal distance over all its assigned part (from the first step) is selected as the compensation coefficient (see Fig.~\ref{fig:bounding}d)---thereby guaranteeing the separation distance $S_{\mathrm{p}}$ in all cases. With increasing density of the keypoints, the compensation coefficients get smaller. 

In our case, the robot compensation values were determined from the model of the robot. For the human, the values were assigned empirically based on the distribution of OpenPose keypoints (Table \ref{table:approx_human}). The human operator was interacting with the robot only with his upper body and the lower body was not taken into account. The resulting bounding spheres are in Fig.~\ref{fig:bounding} and the values are in Table~\ref{table:approx_human}. 


\begin{table}[htb]
\centering
\begin{tabular}{ccccccccc}
\textbf{EE} & \textbf{7} & \textbf{6} & \textbf{5} & \textbf{4} & \textbf{3} & \textbf{2} & \textbf{1} & \textbf{Base} \\
 0.01 & 0.11 & 0.15 & 0.15 & 0.15 & 0.15 & 0.15 & 0.14 & 0.10 \\
& \textbf{Nose} & \textbf{Neck} & \textbf{Eye}  & \textbf{Ear} & \textbf{Arm}  & \textbf{Elbow} & \textbf{Wrist} & \\
& 0.10  & 0.25 & 0.10 & 0.10  & 0.15 & 0.15  & 0.15 & \\
\end{tabular}
\caption{Robot $\mathbf{r}_{\mathrm{compen}}$ and human $\mathbf{h}_{\mathrm{compen}}$ compensation values in meters.}
\label{table:approx_human}
\end{table}

\subsection{Protective separation distance}
\label{subsec:protective_separation_distance}
The \textit{protective separation distance} is the ``shortest permissible distance between any moving hazardous part of the robot system and any human in the collaborative workspace'', $S_p$, and it is described in \cite{ISO/TS15066} by the following formula:

\begin{equation}
    S_p(t_0) = S_{\mathrm{h}} + S_{\mathrm{r}} + S_{\mathrm{s}} + C + Z_{\mathrm{d}} + Z_{\mathrm{r}} \label{eq:protective_separation_distance}
\end{equation}

with
\begin{description}
    \item [$S_{\mathrm{h}}$] contribution to the $S_p(t_0)$ attributable to the operator’s change in location;
    \item [$S_{\mathrm{r}}$] contribution to the $S_p(t_0)$ attributable to the robot system’s reaction time;
    \item [$S_{\mathrm{s}}$] contribution to the $S_p(t_0)$ due to the robot system’s stopping distance;
    \item [$C$] distance that a part of the body can intrude into the sensing field before it is detected;
    \item [$Z_{\mathrm{d}}$] position uncertainty of the operator in the collaborative workspace, as measured by the presence sensing device resulting from the sensing system measurement tolerance;
    \item [$Z_{\mathrm{r}}$] position uncertainty of the robot system from the accuracy of the robot position measurement.
\end{description}

$S_p(t_0)$ can either be calculated dynamically or, as in our case, a fixed value based on worst case situation. Eq. \ref{eq:protective_separation_distance} applies to all personnel in the collaborative workspace and to all moving parts of the robot system. In our case, we calculated the necessary stopping distance based on the maximal robot end-effector speed measured during the robot's unconstrained movement. The contributions marked as $S_i$ are determined using the robot's maximal speed $v_{max}$  multiplied with the appropriate $t_i$, so for example it should be $S_{\mathrm{r}} = t_{\mathrm{r}} \cdot v_{max}$. 
However, we used the average robot speed, $v_{\mathrm{r}}=\frac{v_{\mathrm{max}}-0}{2}$, in our calculations in order to simulate the robot's slowing down during the stopping movement. This is a slight alteration of the very conservative demands of \cite{ISO/TS15066}.

We determined the terms of Eq.~ \ref{eq:protective_separation_distance} as follows:
\begin{description}
    \item [$S_{\mathrm{h}}$] $(t_{\mathrm{r}} + t_{\mathrm{s}}) \cdot v_{\mathrm{h}}$, where $v_{\mathrm{h}}$ is the default human walking speed (1.6 m/s)~\cite{ISO/TS15066}, $t_{\mathrm{r}}$ is the time it took the robot to react to a issued stop status (0.1 s), and $t_{\mathrm{s}}$ the time it took the robot to stop its movement: 0.43 s, thus $1.6\cdot(0.1+0.43) =$ 0.85 m;
    \item [$S_{\mathrm{r}}$] $t_{\mathrm{r}} \cdot v_{\mathrm{max}}$ = $0.1 \cdot 1 = 0.1$ m;
    \item [$S_{\mathrm{s}}$] $t_{\mathrm{s}} \cdot v_{\mathrm{r}}$ = $0.43 \cdot 0.5$ = 0.22 m;
    \item [$C$] the setup did not allow the operator to enter the workspace without being detected: 0 m;
    \item [$Z_{\mathrm{d}}$] see the $h_{\mathrm{compen}}$ values from Subsection~\ref{subsec:bounding_spheres}: 0 m;
    \item [$Z_{\mathrm{r}}$] the LBR iiwa's repeatability value: 0.0001 m.
\end{description}

The time $t_{\mathrm{s}}$ was determined based on measured calculation times (0.005 s) and the maximal deceleration of the robot which was set to 1.5~rad /s$^2$.

Using these values, we can calculate  the $S_p$ as in Eq.~\ref{eq:protective_separation_stop}. 

\begin{equation}
    S_p(t_0) = 0.85 + 0.1 + 0.22 + 0.0001 = 1.17\text{ m}
    \label{eq:protective_separation_stop}
\end{equation}

\subsection{Power and force limiting}
\label{subsec:PFL}
The SSM regime prescribes that the robot stops before contact occurs. In our approach, we also allow the robot to slow down so that it can operate in the PFL regime, see below. We assume the end-effector exerts pressure on a surface area of at least 1~cm $^2$.

We can calculate the maximal relative speed of the system for a transient contact given the surface and the robot weight. For this, we use the formula A.6 from \cite{ISO/TS15066}. This equation also asks for some preliminary calculations, like for example $\mu$, the reduced mass for the two body system of the robot and the human operator. We summarize the calculation here. In order to ascertain absolute safety, we assume the worst case scenario, i.e. an impact in the chest. The values for $m_{\mathrm{h}}$, $p_{\mathrm{max}}$ and $k$ are taken from the appropriate tables in \cite{ISO/TS15066}.

\begin{equation}
    m_{\mathrm{r}} = \frac{M}{2} + m_L = \frac{23.9}{2} + 0
\end{equation}
\begin{equation}
    \mu = \left (\frac{1}{m_{\mathrm{h}}} + \frac{1}{m_{\mathrm{r}}}\right )^{-1} = \left (\frac{1}{40} + \frac{2}{23.9}\right )^{-1}
\end{equation}

\begin{equation}
    v_{rel,max} =  \frac{p_{max} \cdot A}{\sqrt{\mu \cdot K}} = \frac{2.4\times 10^{6} \cdot 1\times 10^{-4}}{\sqrt{\mu \cdot 2.5\times 10^{4}}} = 0.50
\end{equation}

Thus we know that the speed of 0.42 m/s is a conservative speed in order to be in the PFL regime. We determine the distance at which the robot needs to start slowing down to be PFL compliant in the same way as we did with SSM in Eq.~\ref{eq:protective_separation_stop}. However, we take into account only the difference between 1 m/s and 0.42 m/s. The resulting value for $S_p$ is 0.73~m (full to reduced speed). The stopping distance for 0.42 m/s according to the equation would be 0.60~m (reduced to stop). According to \cite{ISO/TS15066}, non-zero energy contact with the human head is not allowed. Thus our final setup forces the robot to stop on the proximity of the human head (see Section~\ref{subsec:scenario5}).

\subsection{Keypoint separation distance representation}
\label{subsec:sep_dist_repr}
The separation distance is represented in a matrix of minimal effective separation distances for every pair of human-robot keypoints that allow to meet the desired protective separation distance for all. This matrix can be set explicitly or it can be a sum of different matrices as in our case.

The resulting separation distance is composed of several components---a $baseline$ and any terms relevant from the safety perspective. The $baseline$ is determined by the experimenter or calculated according to the methodology described together with  Eq.~\ref{eq:protective_separation_distance} in Sec.~\ref{subsec:protective_separation_distance}. We have to evaluate the maximum possible speed and the protective separation distance based on the ``worst cases over the entire course of the application''\cite{ISO/TS15066}. The resulting keypoints $S_{\mathrm{p}}^{ij}$ are added to compensation coefficients based on the bounding spheres $\mathbf{h}_{\mathrm{compen}}$ and $\mathbf{r}_{\mathrm{compen}}$ described already in Sec.\ref{subsec:bounding_spheres}.

This addition leads to the keypoint separation distances $S_{\mathrm{kp}}^{i,j}$ between any two given keypoints $i$, $j$.

\begin{equation}
 \begin{matrix}
  S_{\mathrm{kp}}^{ij} = & h_{\mathrm{compen}}^{i} + & S_{\mathrm{p}}^{ij} & + r_{\mathrm{compen}}^{j}
 \end{matrix} \label{eq:vectors}
\end{equation}

Thus we calculate the keypoint separation distances for each keypoint pair. We show two calculations: (1) According to SSM, the values necessary for a cat. 2 stop from full speed based on the Eq. \ref{eq:protective_separation_stop} with the addition of the compensation values from Table \ref{table:approx_human} according to Eq. \ref{eq:vectors} are shown in  Table~\ref{table:stops}  (left). (2) Combination of SSM and PFL regimes: robot first slows down and then stops only if needed. We add the calculations from Section~\ref{subsec:protective_separation_distance}; the resulting values are in Table~\ref{table:stops} (middle). 
An example is provided in Eq. \ref{eq:reduced_speed} with the nose-end-effector keypoint pair. Reduced speed is triggered at the distance $S_{\mathrm{reduced,kp}}^{i,j}$ that is composed of $S_{\mathrm{full to reduced}}$ per PFL (Section~\ref{subsec:PFL}) and $S_{\mathrm{reduced to stop,kp}}$ per SSM (Section~\ref{subsec:protective_separation_distance}, Table~\ref{table:stops}, last column). 

\begin{equation}
 \begin{matrix}
  S_{\mathrm{reduced,kp}}^{i,j} & = & S_{\mathrm{full to reduced}} & + &  S_{\mathrm{reduced to stop,kp}}^{i,j} \\
  1.44 & = & 0.73 & + & 0.71
 \end{matrix} \label{eq:reduced_speed}
\end{equation}

Because of the shape of the KUKA robot, the values result in similar effective $S_{kp}$; accordingly we list three keypoints from the robot and omit duplicate keypoint-pair values.


\begin{table}[tb]
\centering
\begin{tabular}{ccc|cc|cc}
~ & Stop  &  from & Reduce  & speed & Stop   & from  \\
~ & full & speed & ~  & ~ & reduced  & speed  \\
~ & \textbf{Nose} & \textbf{Wrist} & \textbf{Nose} & \textbf{Wrist} & \textbf{Nose} & \textbf{Wrist}\\
\textbf{End-effector} & 1.28  & 1.33 & 1.44 & 1.49 & 0.71  & 0.76 \\
\textbf{3} & 1.33 & 1.38 & 1.49 & 1.54 & 0.76 & 0.81 \\
\textbf{Base} & 1.28 & 1.33 & 1.44 & 1.49 & 0.71  & 0.76\\
\end{tabular}
\caption{Effective keypoint-pair protective separation distance in meters.}
\label{table:stops}
\end{table}

\section{Results}
\label{sec:results}
The robot performs a mock pick-and-place task; the operator periodically replaces one of the objects, entering the robot workspace. The robot responds appropriately by slowing down or stopping and resumes operation whenever possible. 
The scenarios contrast the standard approach of a zone scanner or safety mat (Sc. 1, 2) with the pairwise distance evaluation between operator and robot keypoints (Sc. 3-5). Some scenarios employ a safe reduced speed per PFL (Sc. 2, 4, 5) and Sc. 5 issues a stop only on human head proximity.
The description of the scenarios in our implementation (Sec.~\ref{subsec:scenario1} -- ~\ref{subsec:scenario5}) is followed by a performance comparison on the mock task (Sec.~\ref{subsec:benchmark}).
All upper body keypoints (see Fig.~\ref{fig:bounding}, right) were considered at all times, but we show only the safety-inducing keypoints in the plots below for clarity.

\subsection{Scenario 1 and 2: Robot base vs. human keypoints}
\label{subsec:scenario1}
In the first two scenarios, the distances between the robot base and the human keypoints were considered. The baseline $S_{\mathrm{p}}$ of 1.17~m (Eq.~\ref{eq:protective_separation_stop}) is extended by compensation coefficients specific to the human keypoint bounding spheres (Sec. \ref{subsec:bounding_spheres}, Table~\ref{table:stops}). In addition, as only the base of the manipulator is considered, the robot's maximum reach of 0.8~m has to be added, giving 1.17+0.8 m, plus keypoint compensations.

In a similar manner, the second scenario approximated the setting with distance-based zones for reduced speed and stopping by using the values from Sec. \ref{subsec:PFL}. A reduced speed zone started at 2.13~m (0.73+0.6+0.8) and stop at 1.40~m (0.6+0.8). The separation distance for slowing down from the maximum velocity was a composition of the necessary distance for slowing down, the necessary distance to stop from the reduced speed, and the robot's reach, see Fig. \ref{fig:bounding}a.


\subsection{Scenario 3 and 4: Robot vs. human keypoints}
\begin{figure}[htb]
	\centering
	\includegraphics[width=\columnwidth]{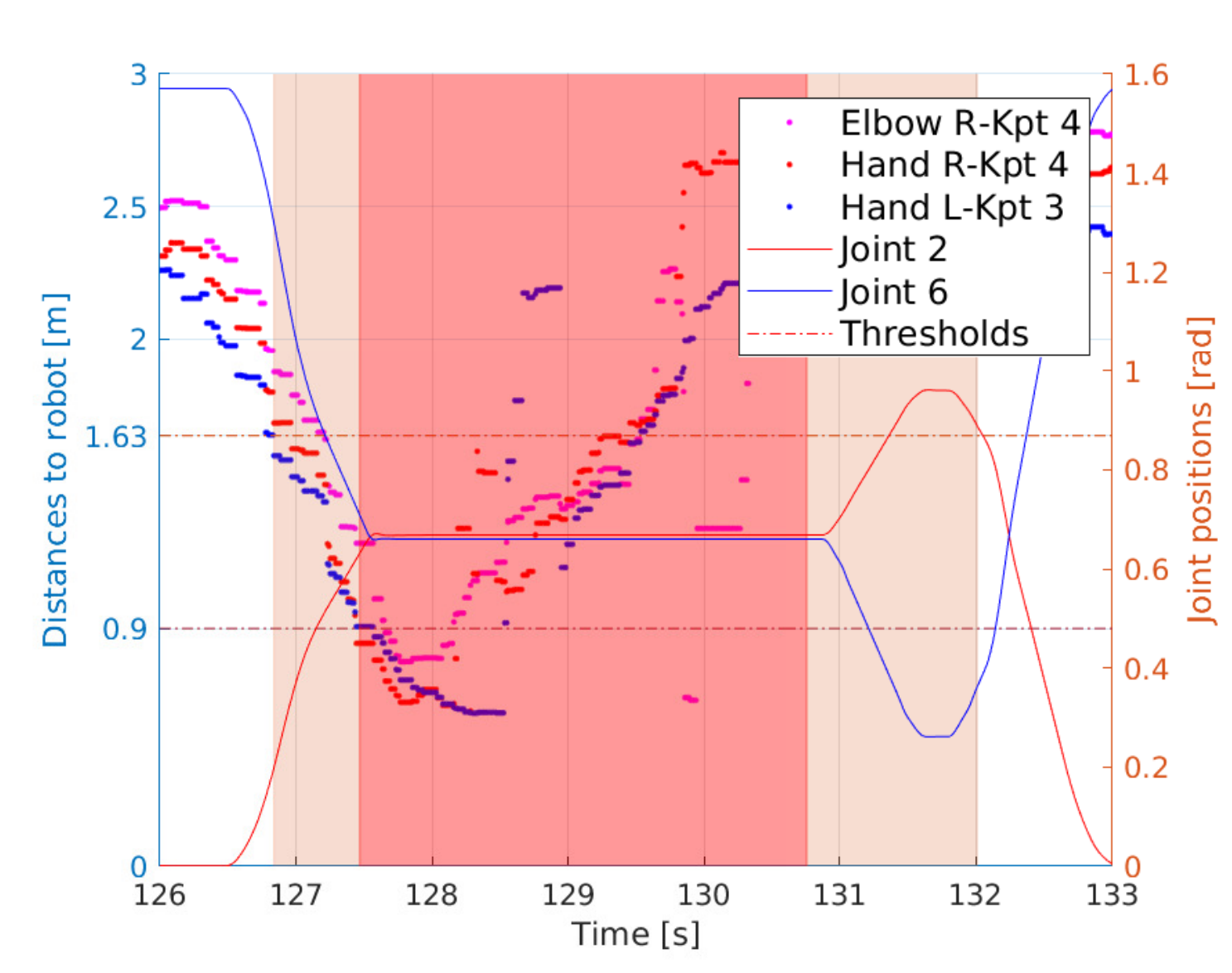}
	\caption{Scenario 4: Reduced speed (light area) or stop (dark) triggered by keypoint distances below threshold. Positions of selected joints showing the slowing down / stopping (continuous lines, right y-axis). Keypoint pair distances triggering the behavior are shown (individual data points, left y-axis). Relevant threshold values: Reduced speed at 1.63~m and the stopping behavior at 0.90 m. These values are based on Eq. \ref{eq:reduced_speed} and the appropriate compensation values from Table \ref{table:approx_human}. }
	\label{fig:scenario4}
\end{figure}

In Scenario 3, we measure keypoint-pair separation distance with respect to the robot's moving parts (namely any joint above joint 3) to stop at  $S_{\mathrm{p}}$ = 1.17 m.
The fourth scenario involved a reduced speed zone (see Sec. \ref{subsec:PFL}).  When a human keypoint got closer than 1.33~m to any of the moving robot keypoints, the robot slowed down. If the human got closer than 0.60 m, the robot stopped.
The behavior of the system is illustrated in Fig.~\ref{fig:scenario4}.

\subsection{Scenario 5: Addition of keypoint discrimination}
\label{subsec:scenario5}
\begin{figure}[htb]
	\centering
	\includegraphics[width=\columnwidth]{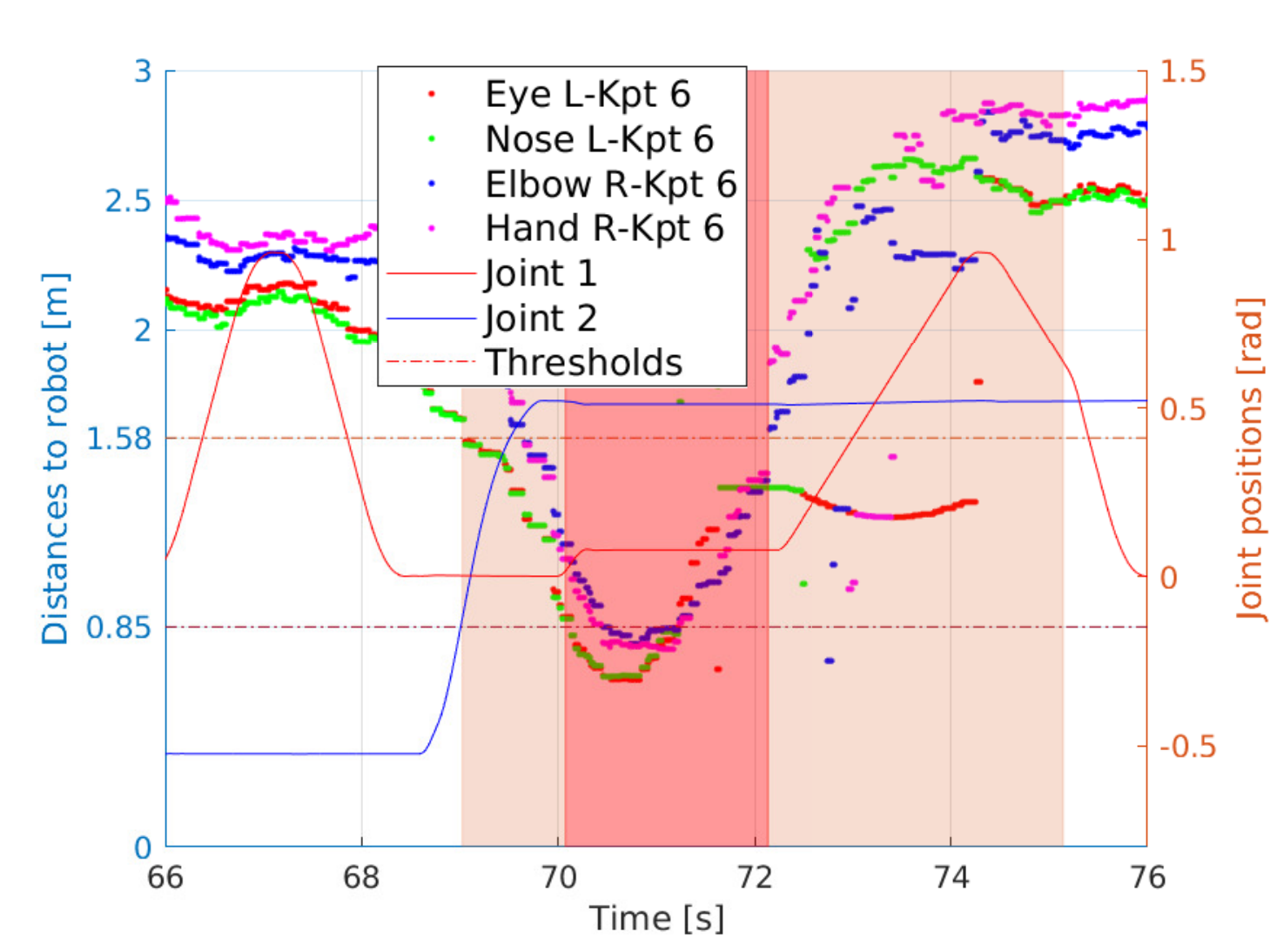}
	\caption{Scenario 5. See also caption of Fig.~\ref{fig:scenario4}. As soon as the first threshold at 1.58~m is met, the robot reacts with slowing down. When the human operator crosses the second threshold at 0.85~m with his head, the robot stops. Thresholds contain the compensation from Sec. \ref{subsec:bounding_spheres}. Notice that the detection of the operator's elbow below the threshold does not trigger a stop but it does lead to a longer reduced speed period.}
	\label{fig:scenario5}
\end{figure}

The last scenario described the case when the robot reacted with a stop only if the human head was closer than 0.60~m to the robot. Otherwise, the robot slows down (keypoint distance below 1.33 m). The behavior is illustrated in Fig.~\ref{fig:scenario5}. Notice that the safety regimes of the robot were triggered by different keypoint pairs than in the case of the previous scenario in Fig.~\ref{fig:scenario4}.

\subsection{Performance in mock task}
\label{subsec:benchmark}
 Here we quantitatively evaluate the performance on the task under the different ``safety regimes'' as described above. The robot performs the task 20~times (measured at one of the two target objects) and the time needed is recorded. As a baseline, we use the unobstructed task at full speed of the robot and reduced speed. The full speed scenario would not comply with collaborative operation; reduced speed at all times would comply, provided the operator head is protected.

The results are shown in Table~\ref{table:speed_results}. Operating the robot in the reduced speed PFL compliant regime, scenarios 4 and 5, outperformed most of the experimental scenarios. The scenarios that take pairwise distances between robot and operator keypoints into account and use two thresholds (scenario 4 and 5) performed better than all other collaborative regimes. The last scenario that stops only for the head keypoints achieves the best performance.

\begin{table}[htb]
\centering
\begin{tabular}{ccccccc}
\textbf{Full sp.} & \textbf{Reduced sp.} & \textbf{Sc. 1}  & \textbf{Sc. 2} & \textbf{Sc. 3} & \textbf{Sc. 4} & \textbf{Sc. 5} \\
154 & 256 & 267 & 254 & 257 & 231 & 228 \\
\end{tabular}
\caption{Task duration for different scenarios in seconds.}
\label{table:speed_results}
\end{table}

\section{Discussion and conclusion}
In this work, we used a robot in a mock collaborative scenario, in which it shares its workspace with a human. The operator's position was perceived with an Intel RealSense RGB-D sensor and human keypoints were extracted using OpenPose. Our paper presents an application of the standard for collaborative robot operation ISO/TS 15066 \cite{ISO/TS15066}. The standard prescribes two collaborative regimes (SSM and PFL). However, to our knowledge, there is no work considering both in a single application. We follow the standard to derive the protective separation distance (per SSM) and calculate the reduced robot velocity (in compliance with PFL constraints) and deploy them in a single framework. We demonstrate this union with an implementation of pairwise keypoint distance monitoring. Compared to classical zone monitoring, the keypoint distance method has higher resolution and constraints robot operation less. Also, keypoints can be treated differently, taking the sensitivity of human body parts or robot keypoints (e.g. sharpe edges) into account---in this way the constraints on collisions (per PFL) can be transformed into separation distances (per SSM).


The operation of this framework was illustrated with a KUKA LBR iiwa robot interacting with a human partner that is perceived by a RGB-D sensor during a mock collaborative task. Contrasting a classical ``stop zone'' from the robot base with the keypoint-based approaches confirmed the potential of the distance monitoring between pairs of keypoints. 

Multiple features could enhance our setup, notably we could add dynamic protective separation distances and occlusion compensation. The current approach monitors only positions and uses the maximum speeds for calculations. Instead, we could monitor  relative speed and dynamically modify the protective separation distance accordingly.

Currently, occlusions could cause a misestimation of the human's keypoint location and thus the distance. Possible compensations and thus future enhancements are to use multiple sensors, compensate for occlusion by creating a human model or filter out the robot body in the scene. With these additions we could also incorporate active evasion of the human instead of our current reactive behavior (see \cite{Flacco2015}). 

RGB-D sensors are not safety-rated yet. The reliability of the current sensors can be improved by combining multiple sensors and fusing the information from them \cite{Flacco2017,Ragaglia2018}. 
However, there is a clear need of safety-rated devices similar to those for zone monitoring that will provide 3D object coordinates and possibly human keypoint extraction: certified products are expected to appear on the market soon.
The availability of such technology would dramatically expand the possibilities of human-robot collaboration in the SSM regime. Furthermore, as illustrated in this work, exploiting the ``keypoint semantics'' (e.g. chest vs. head) can be combined with the safety requirements as per PFL. 




\section*{ACKNOWLEDGMENT}
This work was supported by the Czech Science Foundation, GA17-15697Y (P.S., M.H.); the Technological Agency of the Czech Republic, TJ01000470 (M.T.); the Czech Technical University in Prague, grant No. SGS18/138/OHK3/2T/13 (P.S.); the European Regional Development Fund, ``Research Center for Informatics'' (CZ.02.1.01/0.0/0.0/16\_019/0000765) (P.S.);  ``Robotics for Industry 4.0'' (CZ.02.1.01/0.0/0.0/15 003/0000470) (J.K.B.).  We thank Karla Stepanova for assistance, Zdenek Straka for his previous work \cite{Svarny_SSR_2018}. We are also indebted to Vasek Hlavac, Valentyn Cihala, Libor Wagner, Vladimir Petrik, Vladimir Smutny, and Pavel Krsek from CIIRC for their kind support in using the KUKA robot.

\bibliographystyle{IEEEtran}
\bibliography{SvarnyTesarHoffmann_KukaRealSenseHRI}

\end{document}